\documentclass{amia}
\usepackage{graphicx}
\usepackage[labelfont=bf]{caption}
\usepackage[superscript,nomove]{cite}
\usepackage{color}
\usepackage{amsmath}
\usepackage{ulem} 
\usepackage[bb=dsserif]{mathalpha}
\usepackage[dvipsnames]{xcolor}
\usepackage{bm}
\usepackage{url}
\usepackage{svg}
\usepackage{color-edits}
\addauthor{TC}{blue}

\begin{document}


\title{Backdoor Adjustment of Confounding by Provenance for Robust Text Classification of Multi-institutional Clinical Notes}

\author{Xiruo Ding, MS$^{1}$, Zhecheng Sheng, MS$^{2}$,\\ Meliha Yetişgen, PhD$^{1}$, Serguei Pakhomov, PhD$^{2}$, Trevor Cohen, MBChB, PhD$^{1}$}

\institutes{
    $^1$University of Washington, Seattle, WA, USA\\
    $^2$University of Minnesota, Twin Cities, MN, USA\\
}

\maketitle

\noindent{\bf Abstract}

\textit{Natural Language Processing (NLP) methods have been broadly applied to clinical tasks. Machine learning and deep learning approaches have been used to improve the performance of clinical NLP. However, these approaches require sufficiently large datasets for training, and trained models have been shown to transfer poorly across sites. These issues have led to the promotion of data collection and integration across different institutions for accurate and portable models. However, this can introduce a form of bias called confounding by provenance. When source-specific data distributions differ at deployment, this may harm model performance. To address this issue, we evaluate the utility of backdoor adjustment for text classification in a multi-site dataset of clinical notes annotated for mentions of substance abuse. Using an evaluation framework devised to measure robustness to distributional shifts, we assess the utility of backdoor adjustment. Our results indicate that backdoor adjustment can effectively mitigate for confounding shift.}


\section{Introduction}
Machine learning methods are well-established in clinical natural language processing (NLP) \cite{percha2021modern}, and the recent successes of deep neural network models have generated interest and enthusiasm in their applications in the clinical domain \cite{wu2021deep}. Adequate amounts of diverse data are key to successful training of such models. However, this need for size and diversity presents a considerable challenge for clinical data collection from an individual institute\cite{shellerFederated2020}. Integrating data from multiple institutions is a natural way of increasing dataset size, and also promoting collaboration. Data coming from different sources will benefit model training, but at the same time, may also introduce bias. Recent work in NLP has drawn attention to the potential of confounding variables – variables that influence both the predictors (incoming text) and outcomes (an assigned category) of a text categorization system – to harm model performance at the point of deployment \cite{landeiro2016, landeiro2018}. Essentially, the concern is that when the distribution of category assignment in the training data varies in accordance with the value of a confounding variable (e.g. gender of the author), the model will erroneously learn to make category assignments using words that are associated with this variable (e.g. gender differences in pronoun use)\cite{newman2008}, rather than words that meaningfully inform the categorization task at hand. In our recent work, we have identified \textit{confounding by provenance} as a variant of this confounding effect in which models learn to associate features that indicate the source (provenance) of a component of a multi-institutional dataset with this component's label distribution \cite{howell2021, yue2021}. Confounding by provenance threatens the integrity of machine learning models trained on multi-institutional data sets, and may lead to inaccurate predictions once they are deployed in settings where source-specific distributions of the category of interest differ from training data, with potential to limit adoption of AI models and threaten patient safety.

In this work, we use a backdoor adjustment for text classification method developed by Landeiro and Culotta \cite{landeiro2016, landeiro2018},  which follows a similar form to backdoor adjustment for causal inference, as introduced by Pearl \cite{causality2009}. This method was shown to reduce confounding bias in a set of clinical notes drawn from different services within an institution in our recent work on detection of documented goals-of-care discussions \cite{howell2021}.  A limitation of this prior work, including our own, concerns how texts were represented: as binary unigram vectors. The availability of large language models presents representational alternatives that involve encoding text as continuous vectors (embeddings). Methods such as Bidirectional Encoder Representations from Transformers (BERT)\cite{bert} and Sentence-BERT\cite{sbert2019} generate text embeddings that naturally address the perennial NLP concerns of synonymy (because similar words will produce similar embeddings) and polysemy (because representations of the same word will differ in accordance with local context). One goal of the current research is to evaluate their amenability to backdoor adjustment as a means to address the problem of confounding by provenance, in comparison with binary unigram vectors. In addition, we wished to go beyond our prior evaluation efforts to develop a principled framework for the evaluation of robustness to \textit{confounding shift} \cite{landeiro2018} - a shift in positive class probability given a confounding variable between training and test data - in the context of confounding by provenance.

The main contributions of our work are:
\begin{itemize}
   \item The formal definition of the problem of confounding by provenance (Section \ref{sec:Problem}). 
    \item The development of an evaluation framework for robustness to provenance-related confounding shift (Section \ref{sec:PerturbDist}).
    \item The deployment of this framework in an evaluation of the effectiveness of Landeiro and Culotta's backdoor adjustment method when applied to Sentence-BERT embeddings (Section \ref{sec:sbert}).
\end{itemize}

\section{Problem Definition and Dataset} \label{sec:Problem}
\subsection{Confounding by Provenance} 
In our recent work, we observed confounding by provenance when data from multiple sources were combined together for model development \cite{howell2021, yue2021}. In both of the settings concerned, some level of data integration from different sources was applied. In one case, a transfer learning framework \cite{yue2021} was applied. In the other, the application involved deliberately merging data from different types of clinical notes \cite{howell2021}. Though confounding by provenance was recognized in this work, the papers concerned do not include a formal definition of this phenomenon, nor do they provide framework for the evaluation of robustness to it in the context of confounding shift.

\begin{figure}[ht]
	\centering
	\captionsetup{justification=centering}
	\includegraphics[scale=0.5]{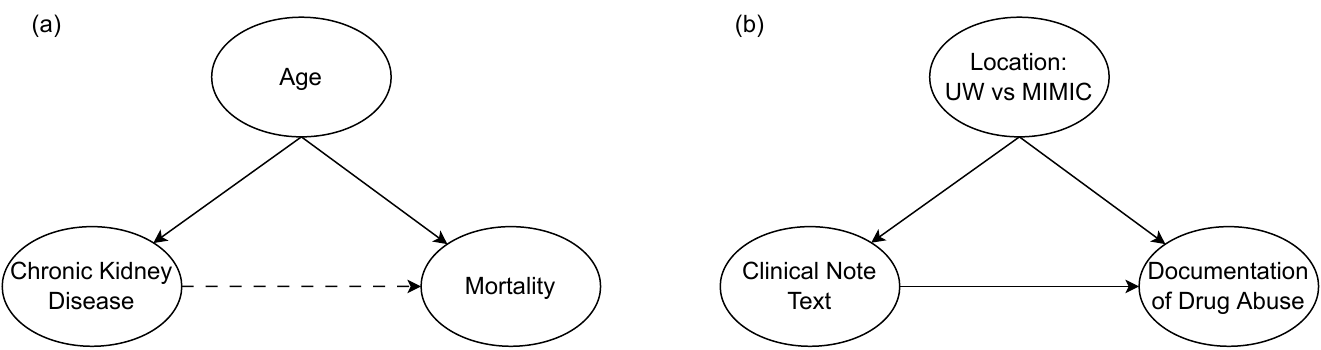}
	\caption{(a) Causal DAG depicting age acts as a confounder in effect of CKD on mortality (b) Non-causal DAG in text classification setting for SHAC dataset, with confounding by provenance of two sources: UW and MIMIC}
	\label{fig:causalDAG}
\end{figure}

Here, we formulate the combination of confounding by provenance and distribution shift as follows. In causal inference, a Directed Acyclic Graph (DAG) provides a convenient way to represent confounding variables \cite{causality2009}. The example in (Figure \ref{fig:causalDAG}(a)) shows a case where the goal is to investigate the effect of chronic kidney disease on mortality. Age is a recognized risk factor to chronic kidney disease \cite{ageCKD2007, ckdFactors2013}, and in the same time it affects mortality rates. Thus, age is considered as a \textit{confounder} in causal inference literature - a variable that influences both the predictor (chronic kidney disease) and the outcome (mortality) of a study.

In the setting of text classification, Landeiro et al. adopted a similar ``non-causal'' DAG for predictive models \cite{landeiro2016, landeiro2018}. Instantiated for confounding by provenance in the context of the dataset used for the current experiments (this is derived from the Social History Annotation Corpus - SHAC \cite{shac2021} and described in Subsection \ref{sec:shac}), text features from clinical notes ($\mathbf{X}$) serve as inputs into classification models (Figure \ref{fig:causalDAG}(b)). Text features will vary by source ($Z$), and both these features and the distribution of positive examples from each source will affect the prediction of drug abuse ($Y$). The distributions of language and labels from two distinct data sources ($Z$) will affect both the text in clinical notes (the predictors) and the frequency (and predicted probability \cite{qu2020}) of documentation of drug abuse (the outcome).

In the context of distribution shift, the problem can be formalized as follows:
\begin{align}
    P_{UW}(X,Y) &\neq P_{MIMIC}(X,Y) \\
    P_{UW,train}(X,Y) &\neq P_{UW,test}(X,Y) \\
    P_{MIMIC,train}(X,Y) &\neq P_{MIMIC,test}(X,Y)
\end{align}

where UW and MIMIC are the two sources of data for the multi-institutional set. The equations indicate that the two sources have different positive class prevalence ($P(Y|Z)$), and that \textit{within} each source this prevalence differs at training and test time. These differences and their implications are discussed in greater detail in Section \ref{sec:PerturbDist}, which describes our evaluation framework in which different degrees of distribution shift are imposed by experimental perturbation.

\subsection{The Social History Annotation Corpus (SHAC)}
\label{sec:shac}
The Social History Annotation Corpus (SHAC) is composed of clinical notes, which have been annotated for detecting social determinants of health using an active learning framework\cite{shac2021, shac2023}. The notes were collected from two sources: clinical notes of chronic pain patients from the University of Washington Medical Center(the \textit{UW Dataset}), and discharge notes of intensive care unit patients from MIMIC-III (the \textit{MIMIC Dataset}). As described in previous work\cite{shac2021}, only the social history sections from these notes were selected (using pattern matching) and retained. These served as our text samples for classification. The SHAC corpus was constructed through an active learning framework. SHAC consists of 4,405 (2,528 from UW and 1,877 from MIMIC) annotated social history sections (original distribution: 70\% train, 10\% development, and 20\% test). All development and test data are randomly sampled. Training samples were 29\% randomly selected and 71\% actively selected. Active learning was used to increase the prevalence of critical risk factors in the annotated training data including positive substance use, unemployment, disability, and homelessness. The annotation was completed by four medical students. 

In this work, we collected all notes from the train, development, and test sets into one pool and resampled from it. For the main outcome, we selected ``drug abuse", among other substance abuse (such as alcohol and tobacco), employment, and living status categories. We made this selection because of the outcome variables concerned, drug abuse had the greatest difference in positive class prevalence across the two sources, and as such presented the best opportunity to explore confounding by provenance.  The original annotation was developed to support a span extraction task, and included more granularity than is required for text categorization, including the extraction of modifiers indicating the status of the documented drug abuse.  For our purpose we only used the Status arguments of ``current'' and ``past'' to construct a positive label for drug abuse. All others were considered as negative.


Documented cases of drug abuse are distributed very differently in the two sources, with a positive class prevalence of 41.1\% for the UW dataset and 19.8\% for the MIMIC dataset (Table \ref{table:summaryN}). These differences likely reflect differences in sampling strategies across the two sites, rather than differences in the prevalence of drug abuse at each location\cite{shac2021}.
\begin{table}[H]
\centering
\caption{Summary of total number of notes and identified drug abuses}
 
\begin{tabular}{c|ccc}
\hline
      & Total Number & Identified Drug Abuse Cases & Positive Rate \\ \hline
UW    & 2528         & 1040                        & 41.1\%        \\
MIMIC & 1877         & 371                         & 19.8\%        \\ \hline
\end{tabular}
\label{table:summaryN}
\end{table}

\section{Methods}

\subsection{Backdoor Adjustment in Text Classification}

Backdoor adjustment is a widely used technique and has been studied across different domains where causal inference is needed\cite{backdoor2001, backdoor2020, backdoorPolitical2020, backdoor2021} (for an illustrative recent example we refer the interested reader to this prospective observational study evaluating COVID-19 vaccine side effects when non-causal paths need to be adjusted \cite{menni2021}). 

According to Pearl (Causality, Equation (3.19) in pg. 80), under the simple setting shown in (Figure \ref{fig:causalDAG}(a)), the confounding effect can be controlled for by the following summation over $Z$ \cite{causality2009}:
\begin{align} \label{formula:backdoorAdj}
    P(y|\hat{x}) = \sum_{z}P(y|x,z)P(z)
\end{align}


Landeiro et al. proposed a similar adjustment for text classification \cite{landeiro2018}. At the point of prediction when the true confounding variable is unavailable, the adjusted estimated (in change of adjusted) causal effect can be calculated as:
\begin{align}
    P(y|x) = \sum_{z=c}^{C}P(y|x,z_c)P(z_c)
\end{align}

Essentially, the prediction probability is the sum of the probabilities across 
all possible $Z$ values (for all known categories,$1,2,...,C$). This estimate is used in the absence of knowledge of the true $Z$, in order to compute $P(y|x,z_c)$. Under the logistic regression framework, for any potential $Z=c$, the logit for a single case $i$ is given by:
\begin{align}
   logit(\hat{y_{ic}}) =\beta_0 + \boldsymbol{\beta_1 X_i} + \boldsymbol{\beta_2 z_{ic}} + \epsilon_i \label{formula:LRwBackdoor}
\end{align}
In our experiments on the SHAC dataset, $logit(\hat{y_{ic}})$ is the estimated logit that example $i$ belongs to $c$ (UW or MIMIC), $\boldsymbol{X_i}$ represents the text encoding features (binary unigram vectors or Sentence-BERT embeddings). $P(z_c)$ is empirically inferred from the training set, using frequencies for each category:
\begin{align}
    P(z_c) = \frac{\sum_{i\in D_{train}} \mathbb{1}(y_{i}=c)}{|D_{train}|}  \label{formula:pz}
\end{align}
At this point, we can combine Equations \ref{formula:LRwBackdoor} and Equation \ref{formula:pz} with  Equation \ref{formula:backdoorAdj} to estimate the probability of documented drug abuse for any example $i$ in the testing set.

In our experimental setting for the SHAC dataset, we only control one confounding variable: the source, and it is binary ($z \in \{UW ,MIMIC\}$). It is encoded using a modified one-hot encoding. For example for each instance, we concatenate ($v \times \mathbb{1}(UW)$, $v \times \mathbb{1}(MIMIC)$) as additional features to the vector representing the text. Here, $v$ is a hyperparameter controlling how much to constrain emphasis on the text-derived features, $\boldsymbol{X}$ (relative to $Z$). $v=1$ corresponds to one-hot encoding. Other settings of $v=10$ and $v=100$ were also tested in our experiments.

\subsection{Perturbation of Distributions} \label{sec:PerturbDist}
To support the evaluation of robustness to confounding shift, we designed a perturbation framework for distributions so that different degrees of shift could be simulated by sampling. For the current experiments with the SHAC dataset, we have a binary outcome (documented drug abuse vs. no documented drug abuse) and binary provenance confounder (UW vs. MIMIC). For problems meeting these constraints, the following probabilities govern the distribution of $P(Y,Z)$:
\begin{align}
& p_{train}(y=1|z=0) \label{constraint:1}\\
& p_{test}(y=1|z=0) \label{constraint:2}\\
& p_{train}(y=1|z=1) \label{constraint:3}\\
& p_{test}(y=1|z=1) \label{constraint:4}\\
& p_{train}(y=1) = p_{test}(y=1) = Const_y \label{constraint:y} \\
& p_{train}(z=1) = p_{test}(z=1) = Const_z  \label{constraint:z}
\end{align}

These probabilities determine the source-specific class distributions at training and test time. Of these constraints, (\ref{constraint:y}) and (\ref{constraint:z}) are held constant across experiments while the others are varied. Furthermore, we introduced two auxiliary helper variables $\alpha_{train}$ and $\alpha_{test}$, which measure the train and test set ratios between positive class prevalence in data from each site:
\begin{align}
    &\alpha_{train} = \frac{p_{train}(y=1|z=1)}{p_{train}(y=1|z=0)} \\
    &\alpha_{test} = \frac{p_{test}(y=1|z=1)}{p_{test}(y=1|z=0)} \label{constraint:alphaTest}
\end{align}

Now, given (\ref{constraint:1}), (\ref{constraint:3}), (\ref{constraint:z}), (\ref{constraint:alphaTest}), we can obtain the remaining probabilities for (\ref{constraint:2}), (\ref{constraint:4}), (\ref{constraint:y}). Intuitively, this means all that is needed to set up an evaluation of robustness to confounding shift is to control the training distributions and set a positive ratio between sources for the testing set. This simulates a setting where, in general, we only know our training set, and different degrees of shift are simulated.

We proceed to describe our experiments in detail. The $Z$ variable was assigned as 0 for UW, 1 for MIMIC. We fixed the training set size at 2,000 and testing set size at 500, so that for all settings we introduced confounding shift by undersampling and no instance was drawn more than once. We first sampled our training set to have positive rates similar to those shown in Table \ref{table:summaryN}: 
\begin{align*}
    p_{train}(y=1|z=UW)=0.5\\
    p_{train}(y=1|z=MIMIC)=0.2
\end{align*}

This corresponds to $\alpha_{train}=0.2/0.5=0.4$. Next, $p_{train}(z=1) = p_{test}(z=1) = Const_z$ was drawn from range 0.1 to 0.9 (inclusive) at a step size of 0.05, $\alpha_{test}$ from range 0 to 10 (inclusive) at a step size of 0.05. Excluding extreme combination settings where training and testing set sizes could not be successfully drawn without replacement left with 1,287 valid settings. Each setting represents a different amount of provenance-related confounding shift. Shift toward increased representation of positive examples from MIMIC in the test set will result in higher $\alpha$ values. Shift toward increased representation of positive examples from UW in this set will result in lower $\alpha$ values.

\subsection{Discrete Text Representation Using Binary Unigram Vectors}
As a baseline, binary unigram vectors (i.e. one-hot vectors with a coordinate corresponding to each word in the vocabulary) were used following the work of Landeiro et al. \cite{landeiro2016, landeiro2018} This is a constrained case of an n-gram representation ($n$=1), where only one word is considered when constructing the dictionary. Furthermore, only a binary indicator (0 for absent, 1 for present) of each word is recorded into the final vector for a given document.

\subsection{Continuous Text Representation Using Sentence-BERT} \label{sec:sbert}
Proposed in 2019 by Reimers et al., Sentence-BERT is a BERT-based framework \cite{devlin2018} with modifications in training using siamese and triplet network structures to reduce the distance between vector representations of sentence with similar meaning \cite{sbert2019}. In addition to improving performance on sentence similarity benchmarks, this reduces the computing capabilities required for larger BERT models (such as BERT base) when comparing sentence similarities, especially for long sentences. Several pretrained Sentence-BERT models have been released since then, with varying neural network depths and training corpus. We used the \texttt{all-MiniLM-L6-v2} version, which with only 6 transformer layers provides a good balance between performance and computational efficiency. This pretrained model is publicly available from the HuggingFace repository \footnote{\url{https://huggingface.co/sentence-transformers/all-MiniLM-L6-v2}}. Each document was provided as input to the model and the output, a 384-dimensional dense vector then served as a continuous document representation.

\subsection{Evaluation Setup}
With perturbed distributions, we applied backdoor adjustment for text classification using the logistic regression, using (\ref{formula:LRwBackdoor}). A ``vanilla'' version of logistic regression without the provenance confounder variable $Z$ was was also fit using the same settings as a baseline:
\begin{align}
    logit(y_{ic}) =\beta_0 + \boldsymbol{\beta_1 x_i} + \epsilon_i
\end{align}

Each experiment was repeated five times, with a different random seed for subsampling instances to generate train/test splits with confounding shift. On the testing set, we report the mean and standard deviation of Area Under the Precision-Recall Curve (AUPRC) values for each setting. AUPRC was chosen for its better discriminant ability in rare-case scenarios \cite{auprc2015}.

\section{Results}
\begin{figure}[H]
	\centering
	\captionsetup{justification=centering}
	\includegraphics[scale=0.4]{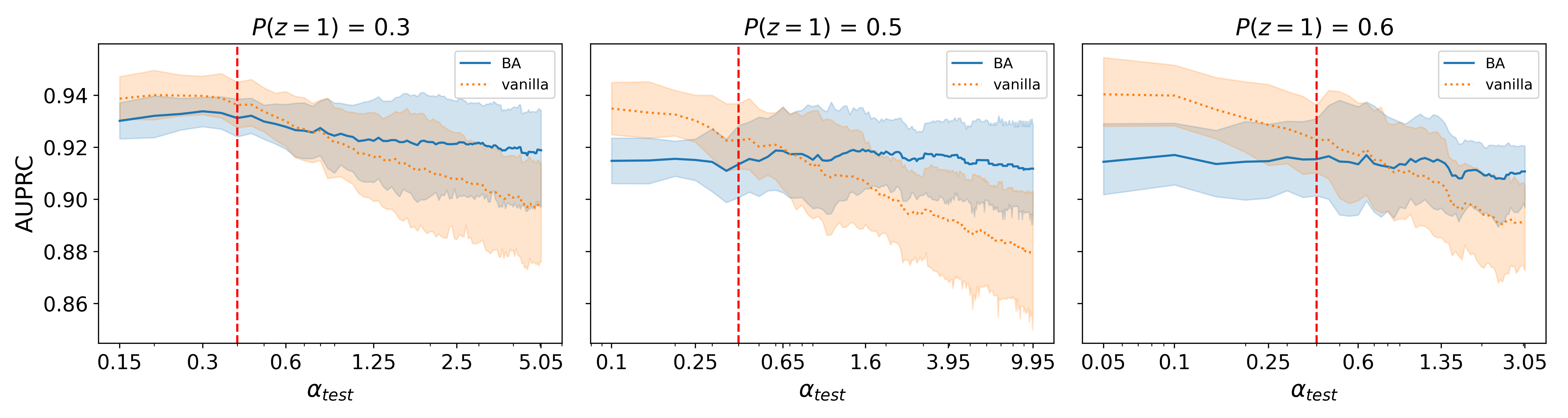}
	\caption{AUPRC performance with binary unigram representations ($log10$ scale for x). $v=10$. Vertical red dashed line in each plot represents $\alpha_{train}=0.4$, where $\alpha_{test}$ matches the training set and distribution difference is minimal. Shaded areas represents 95\% CI for 5 random runs. BA: backdoor adjustment text classification logistic regression. vanilla: simple logistic regression without provenance confounders.}
	\label{fig:result:BU}
\end{figure}

\begin{figure}[H]
	\centering
	\captionsetup{justification=centering}
	\includegraphics[scale=0.4]{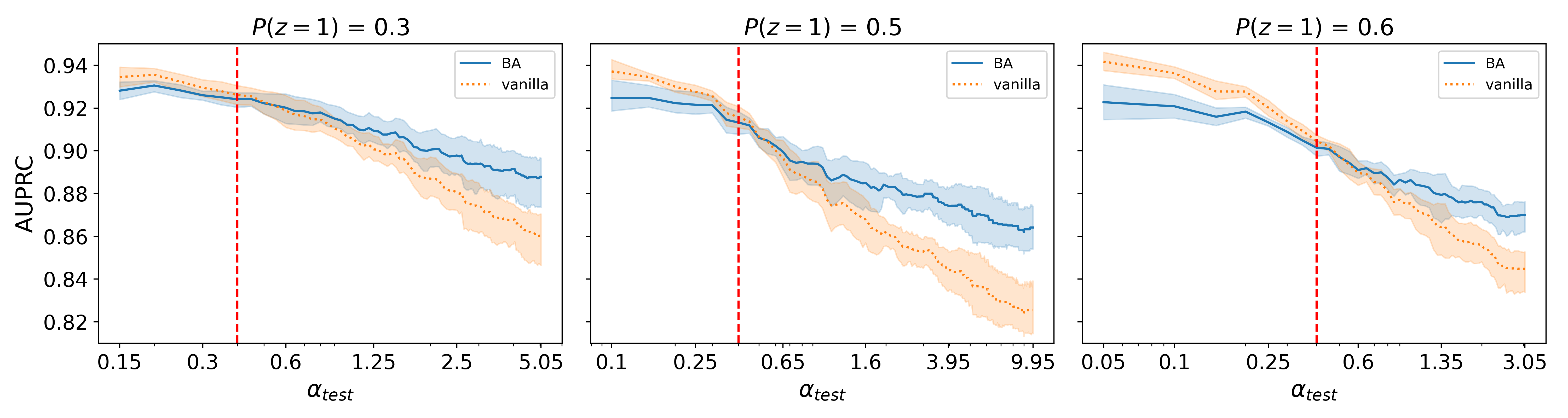}
	\caption{AUPRC performance with Sentence-BERT representations ($log10$ scale for x). $v=10$. Vertical red dashed line in each plot represents $\alpha_{train}=0.4$, where $\alpha_{test}$ matches the training set and distribution difference is minimal. Shaded areas represents 95\% CI for 5 random runs. BA: backdoor adjustment text classification logistic regression. vanilla: simple logistic regression without provenance confounders.}
	\label{fig:result:SBERT}
\end{figure}

Figure \ref{fig:result:BU} shows results for experiments using binary unigrams as text representations. Figure \ref{fig:result:SBERT} shows results for experiments using Sentence-BERT embeddings as text representations. In both cases results for each $P(z=MIMIC)$ setting are shown separately, simulating different mixture ratios of the two data sources. For example, when $P(z=MIMIC)=0.2$, according to (\ref{constraint:z}), for both training and testing set 20\% of data come from UW and the remaining 80\% from MIMIC. For the sake of conciseness, and given the subtle differences between results when varying $P(z=MIMIC)$, only a few settings over the span are shown. 

The figures can be interpreted as follows. The $y$ axis shows model performance on the testing set, measured using the AUPRC. The $x$ axis shows the $\alpha$ value of the test set, which indicates the ratio between the positive class prevalence with $z=1$ (MIMIC) and that with $z=0$ (UW). The $\alpha_{test}$ across settings could not be aligned (as shown from different x-axis ranges) because for some experimental combinations of probabilities, there is no valid way of drawing samples satisfying that constraint. As this ratio moves further from that of the training set ($\alpha_{train}$), indicated by the dashed vertical line, the degree of confounding shift increases. Therefore, a curve that does not drop precipitously while starting from the dashed vertical line and moving to the right on the x-axis indicates robustness to confounding shift in this direction. This is the case with plots of backdoor-adjusted (BA) models (blue lines) only, indicating that this approach is effective at mitigating for confounding shift involving an increase in the proportion of positive examples from MIMIC at test time with both discrete (unigram) and distributed (Sentence-BERT) text representations. However, increasing the proportion of positive examples from the UW site leads to worse performance with backdoor adjustment than with the baseline (``vanilla'') models.

For models using binary unigram representations, BA models lead to much more stable AUPRC results in range of 0.90-0.94 across $P(z=1)$ (MIMIC) of 0.3, 0.5, and 0.6. In comparison, the baseline models show a wider range of 0.87-0.95 in performance. The slopes of the lines in Figure \ref{fig:result:BU} for the two models also indicate the robustness of BA models to confounding shift. For each setting of $P(z=1)$, when $\alpha_{test}$ is small (to the left side of the red dashed line for $\alpha_{train}$ marker indicating an increase in the number of positive examples drawn from the UW site), the ``vanilla'' model performs better than BA with an increase of AUPRC of 0.01-0.02. When moving further to the right side region, the ``vanilla'' model's performance drops and quickly falls beneath that of the BA model (a difference of 0.02-0.04).

Overall, Sentence-BERT embeddings (Figure \ref{fig:result:SBERT}) lead to AUPRCs on the testing set in the range 0.82-0.94. Similar to models using binary unigrams, models using Sentence-BERT embeddings also show different performance according to $\alpha_{test}$ values. In general, when $\alpha_{test}$ is small (to the left side of the red dashed line), the ``vanilla'' models outperform BA models by 0.01-0.02 AUPRC units. To the right of the graph, where $\alpha_{test}$ is large and the proportion of positive examples drawn from MIMIC is high, ``vanilla'' model performance drops. At the rightmost point, BA models performs better by 0.04-0.06 units of AUPRC, where the proportion of positive examples from MIMIC in the testing set is high. Overall, BA models show gentler slopes than baseline models, suggesting robustness provided by backdoor adjustment.

When comparing binary unigram vectors and Sentence-BERT embeddings, in setting of $P(z=MIMIC)=0.5$, Sentence-BERT BA model performance is in the range 0.87-0.93, while binary unigram representations result in an AUPRC of 0.90-0.94. The best performance at optimal $\alpha$ is obtained using Sentence-BERT representations, which is consistent with the general trend in NLP of better performance with deep learning derived distributed representations. However, while there are observable effects relative to the unadjusted baseline, the Sentence-BERT representations appear to be less conducive to backdoor adjustment.  When moving to the right-side region of the vertical red dashed lines (the more $\alpha_{test}$ increases from 0.4), Sentence-BERT models show more reduction in performance, indicating sensitivity to confounding shift. With unigram representations, the slopes of the blue lines (standing for BA models) in Figure \ref{fig:result:BU} are in general much gentler than those in Figure \ref{fig:result:SBERT}, suggesting more robustness to BA models when using binary unigram vectors. When $\alpha_{test}$ is small, especially less than 0.4 (to the left-side region of the red dashed lines), both embedding methods do not help the BA model's performance. Here, the  ``vanilla'' model outperforms it. All of the above observations hold for other settings of $P(z=MIMIC)=0.5, 0.6$. We also performed simulations using different $v$ values: 1, 100 (not included in the paper), and those results remain similar, except that both $v=1$ and $v=100$ lead to a higher variance across repeated experiments. This suggests that $v=10$ provides a good balance between stability and adjustment effect.

\section{Discussion}
In this work, we presented the problem of confounding by provenance, devised a framework for evaluation of robustness to it, and applied this framework to evaluate a method to address it (backdoor adjustment) using the SHAC dataset. To represent text for regression modeling, we used two approaches: binary unigram vectors and Sentence-BERT embeddings. Our findings indicate that the effects of backdoor adjustment vary with this representational choice, as well as with the \textit{direction} of confounding shift (with this direction  determined by the value of $\alpha_{test}$). With a direction of provenance shift promoting higher proportions of MIMIC-derived positive examples in the test set, models with Sentence-BERT embeddings usually perform better in terms of AUPRC (for $P(z=MIMIC)=0.5, 0.6$). However, in the opposite direction of provenance shift promoting higher proportions of UW-derived positive examples in the test set, models with binary unigram vectors slowly drop in performance and at very high $\alpha_{test}$ they outperform models using Sentence-BERT embeddings. Provenance-related confounding shift can happen in both directions (promoting the number of positive test set examples drawn from one source over those from another). Our results suggest that the utility of adjusting for these shifts using the backdoor adjustment methods may vary depending on whether the shifts are toward (to the right of the red dashed lines in figures) or away from (to the left of the red dashed lines in figures) the source with lower prevalence in the training set (MIMIC).

In terms of the relationship between choice of representation and robustness, results from Figure \ref{fig:result:BU} and Figure \ref{fig:result:SBERT} show that models with binary unigram vectors usually have gentler slopes, suggesting stronger robustness to provenance shift in both directions. This is true even for simple logistic regression models without any adjustment for confounders (when comparing orange lines in Figure \ref{fig:result:BU} and Figure \ref{fig:result:SBERT}).

In general, results show that backdoor adjustment for text classification is appropriate method to mitigate for provenance-related confounding shift, and can provide the models with robustness to this shift. The effect of the adjustment, however, varies with different modeling choices, including the section of hyperparameter $v$ for one-hot encoding, text embedding choices, penalization for logistic regression, degree of penalization. $v$ and degree of L2 penalization for logistic regression have modest effects on adjustment, so not all results are shown in this paper.

While these results make a foundational contribution to the study of remediation of confounding by provenance, both our evaluation framework for robustness to it and the methods developed to address this problem have broader implications. They can also be applied in the context of other potential sources of bias by partitioning data sets with this variable, rather than by provenance. For example, one might partition a data set of patient-generated language by patient ethnicity, and use backdoor adjustment to ensure that dialect differences are not being used as spurious cues to predict some unrelated outcome. In this way, both our framework for evaluation and the methods under consideration have the potential to be applied to address other types of bias also.

\section{Limitations}
In this work, we utilized fixed training source-specific positive class rates (0.5 for UW, 0.2 for MIMIC) that are close to those of the full SHAC dataset. Since the positive rates are imbalanced for \textit{training} the models, as future work, we will evaluate the BA method under different settings during training and check its validity when imposing different degrees of provenance shift. This will allow us to assess the relationship between source-specific class distributions at training time and the utility of backdoor adjustment. It is also noted that, from the figures, the densities of points are not uniform in log scale across different $\alpha$ values, due to the fact that we performed uniform sampling of $\alpha$ in the linear space. To avoid potential bias on interpretation, as future work we will update this sampling strategy for a more balanced distribution.

In terms of text feature extractions, we only tested binary unigram vectors as previously used by others in work with backdoor adjustment for text classification \cite{landeiro2016,landeiro2018,howell2021}. Normalized counts of n-grams could be another option for retaining more information. While our results did not show clear benefits for using Sentence-BERT embeddings across all ranges of confounding shift, this may be due to the model capacity limitation of regularized logistic regression. Other statistical and machine learning models, SVM \cite{svm}, XGBoost\cite{xgboost} could be explored under backdoor adjustment. Moreover, according to Sentence-BERT model benchmarks\footnote{\url{https://www.sbert.net/docs/pretrained_models.html}}, the pretrained model ``all-MiniLM-L6-v2'' we used in the paper is not the one with the best performance. Other pretrained models, such as ``all-MiniLM-L12-v2'' and ``all-mpnet-base-v2'', are good candidates for next steps. Finally, to better utilize embeddings generated from large language models, deep neural networks could be used for text classification, in addition to their use for representational purposes. The application of backdoor adjustment while fine-tuning a deep learning model for text categorization remains an interesting direction for future work, though we anticipate many methodological details remain to be resolved.

\section{Conclusion}
In this work, we evaluated the utility of backdoor adjustment as a mean to address confounding by provenance for text categorization. Our result indicate that given the imbalanced source-specific class distributions in our training set, models with the backdoor adjustment generate more stable results than those without it, with both unigram and deep learning derived text representations. Models using binary unigram vectors as input features with adjustment show strong robustness to provenance shift, though this only leads to advantages in performance over a baseline model when the shift is in the direction of the minority source in the training set, raising interesting questions about the scope of applicability of this method. Further work is required to determine its validity under different training set distributions.

\section{Acknowledgements}
This work was supported by U.S. National Library of Medicine Grant (R01LM014056). We owe thanks to Heidi Christensen and Maria Koutsombogera, whose reviews of (Guo et al. 2021) reinforced the pervasiveness of the confounding effect we had recognized.

\makeatletter
\renewcommand{\@biblabel}[1]{\hfill #1.}
\renewcommand\refname{	
	\vspace{-2em}
	\begin{center}
		\text{References}
	\end{center}}
\makeatother

\bibliographystyle{unsrt}

\end{document}